\RequirePackage{silence}
\documentclass{tudelft-report}

\usepackage{caption,booktabs,tabularx}
\usepackage[utf8]{inputenc}
\usepackage{natbib}
\usepackage{longtable}
\usepackage{hyperref}
\usepackage{pdfpages}
\usepackage{subfiles}
\usepackage{changes}
\newcommand{\theTitle}{Text-based classification of interviews for mental health -
juxtaposing the state of the art}
\newcommand{\theSubTitle}{}

\newcommand{\theAuthor}{Joppe V. Wouts}

\newcommand{\theStudentID}{11288140}

\newcommand{\theSupervisor}{dhr. dr. S. van Splunter \\[1em] Drs. A. E. Voppel (UMCG)} 

\newcommand{\theInstitute}{
Informatics Institute\\ 
Faculty of Science\\
University of Amsterdam\\
Science Park 904 \\ 
1098 XH Amsterdam 
}

\newcommand{\theDate}{Jun 26th, 2020}

\begin{document}

\frontmatter




\includepdf{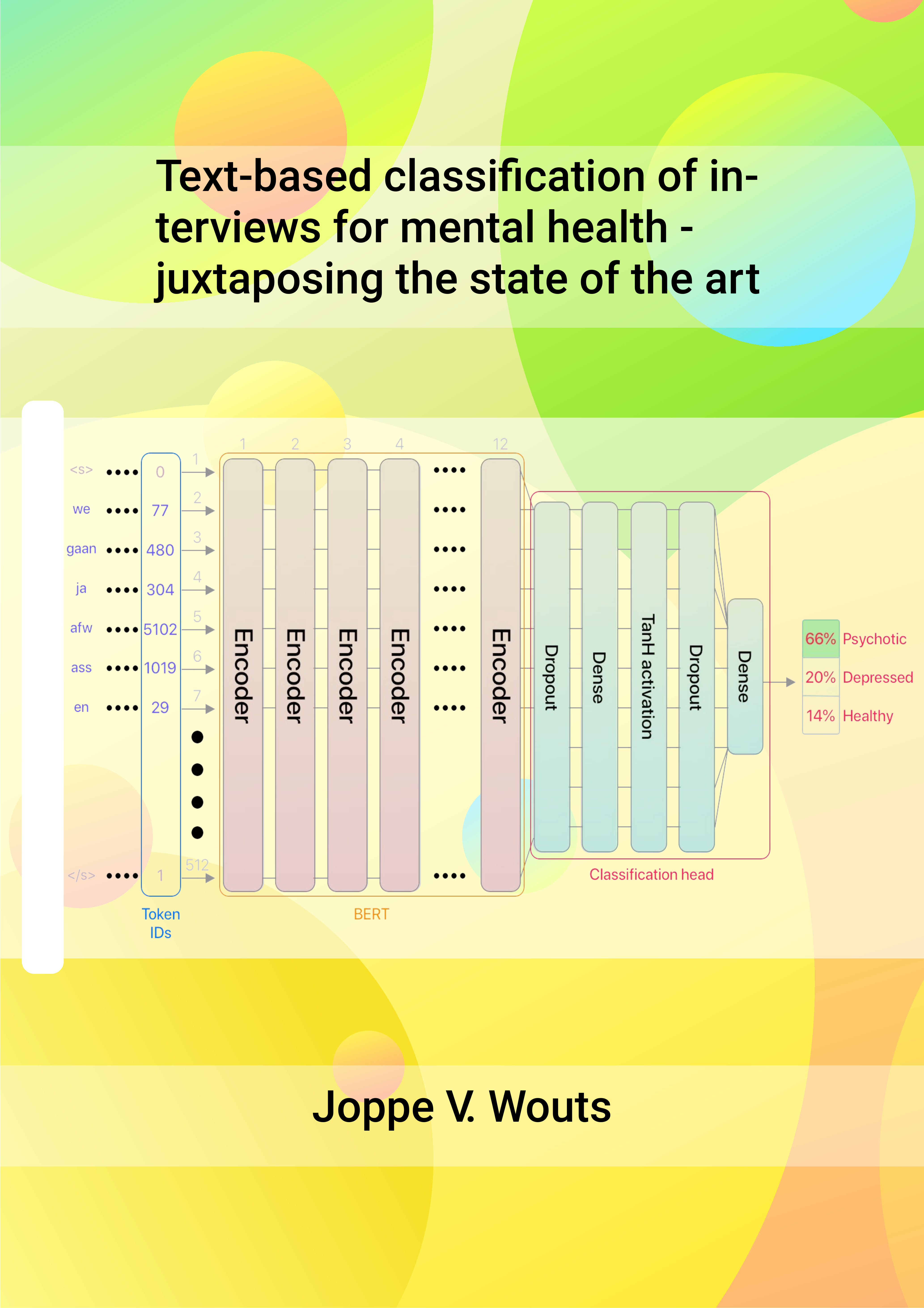}

\begin{titlepage}
\pagestyle{empty}


    
 
        
               
                  
                 
                  
                  
   
    




\begin{center}

\vspace{2.5cm}

\begin{Huge}
\theTitle
\end{Huge} \\

\vspace{0.5 cm}

\begin{Large}
\theSubTitle
\end{Large}

\vspace{1.5cm}

\theAuthor\\
\theStudentID

\vspace{1.5cm}

Bachelor thesis\\
Credits: 18 EC

\vspace{0.5cm}

Bachelor \textit{Kunstmatige Intelligentie} \\
\vspace{0.25cm}
\includegraphics[width=0.075\paperwidth]{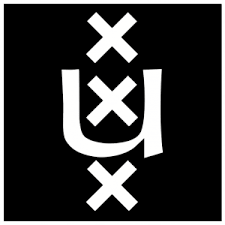} \\
\vspace{0.1cm}

University of Amsterdam\\
Faculty of Science\\
Science Park 904\\
1098 XH Amsterdam

\vspace{2cm}

\emph{Supervisor}\\

\theSupervisor

\vspace{0.25cm}

\theInstitute

\vspace{1.0cm}

\theDate

\end{center}

\end{titlepage}

\tableofcontents

\mainmatter

\chapter*{Abstract}
\setheader{Abstract}

Currently, the state of the art for classification of psychiatric illness is based on audio-based classification. This thesis aims to design and evaluate a state of the art text classification network on this challenge. The hypothesis is that a well designed text-based approach poses a strong competition against the state-of-the-art audio based approaches. Dutch natural language models are being limited by the scarcity of pre-trained monolingual NLP models, as a result Dutch natural language models have a low capture of long range semantic dependencies over sentences. For this issue, this thesis presents belabBERT, a new Dutch language model extending the RoBERTa\cite{liu2019roberta} architecture. belabBERT is trained on a large Dutch corpus (+32GB) of web crawled texts. After this thesis evaluates the strength of text-based classification, a brief exploration is done, extending the framework to a hybrid text- and audio-based classification. The goal of this hybrid framework is to show the principle of hybridisation with a very basic audio-classification network. The overall goal is to create the foundations for a hybrid psychiatric illness classification, by proving that the new text-based classification is already a strong stand-alone solution.

Summarising, the main points of this thesis are

\begin{enumerate}
    \item As the performance of our text based classification network belabBERT outperforms the current state-of-the-art audio classification networks performance reported in literature, as described in section \ref{chap:five}, we can confirm our main hypothesis that a well designed text-based approach poses a strong competition against the state-of-the- art audio based approaches for the classification of psychiatric illness. 
    \item We have shown that belabBERT outperforms the current best text classification network RobBERT. The model of belabBERT is not restricted to this application domain, but generalisable to domains that depend on the capture of long range semantic dependencies over sentences in a Dutch corpus.
    \item We have shown that extending our model to a hybrid model has potential, as performance increased even when adding a simple audio classification network

\end{enumerate}

\chapter*{Acknowledgements}
\setheader{Acknowledgements}

This work was carried out on the Dutch national e-infrastructure with the support of SURF Cooperative. \\

I would also like to thank my supervisors dhr. dr. S. van Splunter and Drs. A. E. Voppel for the extensive discussions and suggestions.

\chapter{Introduction}

Over the last decade psychiatric illnesses have become increasingly prevalent. This has coincided with a problematic trend, which is characterized as a mental health crisis, where according to a Lancet Commission report the worldwide "quality of mental health services is routinely worse than the quality of those for physical health"\cite{patel_lancet_2018}. 

The diagnosis of these illnesses is challenging, as it currently solely relies on subjective reporting\cite{thibaut_controversies_2018}. Accurate diagnosis of psychiatric illnesses remains difficult even for experienced psychiatrists, but even more so for non-specialists such as general physicians or social workers \cite{su_change_2011}.  The latter group of caregivers could form a valuable part of the solution if they were able to accurately assess the presence of these disorders in a patient. 

A potential solution is the use of bio-markers to provide reproducible information on the classification of psychiatric disorders and function as a diagnostic indicator. Analysis of spoken language can provide such a marker. \cite{de_boer_anomalies_2020} \cite{trevino_phonologically-based_2011} Recent technological advances have paved the way for real-time automated speech and language analysis, with state-of-the-art sentiment models reaching 96.21                   \% classification accuracy based on textual data\cite{yang_xlnet_2020}. Speech parameters reflect important brain functions such as motor speed which represents articulation, as well as cognitive functions which are responsible for the correct use of grammar, vocabulary scope, etc. Modern audio analysis can easily extract a variety of low level features which are relevant to different aspects of brain functioning \cite{eyben_opensmile_2010}. Recent research also suggests linguistic and semantic analysis of speech can detect presence of depression, psychosis and mania with >90\% accuracy \cite{de_boer_clinical_2018}. Moreover, other research groups were able to classify post-traumatic stress disorder (PTSD) with an accuracy rate of 89.1\% based on speech markers in audio recordings \cite{marmar_speech-based_2019}.  Language and speech analysis is thus a promising approach to assess a variety of psychiatric disorders etc.

\section{Data description}

A total of 339 participants, of which were 170 patients with a schizophrenia spectrum disorder, 22 diagnosed with depression and 147 healthy controls, were interviewed by a research group of the University Medical Center Utrecht. The interview questions were designed to elicit semi-free speech about general experiences. The interviewers were trained to avoid health related topics in order to  make produced language by the participants more generalisable irrespective of diagnosis or absence thereof.
The raw, digitally recorded audio from the interview was normalized to an average sound pressure level of 60db. 
The openSMILE audio processing framework\cite{eyben_opensmile_2010} \cite{eyben2015geneva} was used to extract 94 speech parameters for each audio file a list of which can be found in table \ref{tab:parameters}.
A subset of each audio file was manually transcribed according to the CHAT \cite{macwhinney_wagner_2010} transcription format by trained transcribers.

\section{Aim of thesis}

Currently, the state of the art for classification of psychiatric illness is based on audio-based classification. This thesis aims to design and evaluate a state of the art text classification network on this challenge. The hypothesis is that a well designed text-based approach poses a strong competition against the state-of-the-art audio based approaches. Dutch natural language models are being limited by the scarcity of pre-trained monolingual NLP models, as a result Dutch natural language models have a low capture of long range semantic dependencies over sentences. For this issue, this thesis presents belabBERT, a new Dutch language model extending the RoBERTa\cite{liu2019roberta} architecture. belabBERT is trained on a large Dutch corpus (+32GB) of web crawled texts. After this thesis evaluates the strength of text-based classification, a brief exploration is done, extending the framework to a hybrid text- and audio-based classification. The goal of this hybrid framework is to show the principle of hybridisation with a very basic audio-classification network.
The overall goal is to create the foundations for a hybrid psychiatric illness classification, by proving that the new text-based classification is already a strong stand-alone solution.

\chapter{Related work}

In this section we explore text and audio analysis techniques suitable for our text classification network and our text- audio hybrid network. The final subsection presents an approach for the hybrid network.

\section{Text analysis}

In the field of text analysis there is a huge variety of approaches ranging from finding characterizing patterns in the syntactical representation of text by tagging parts-of-speech, to representing words as mathematical objects which together form a semantic space, with the latter approach having a rapid rise in various linguistic problems. In a meta-analysis of eighteen studies in which semantic space models are used in psychiatry and neurology \cite{de_boer_clinical_2018} draw the conclusion that analyzing full sentences is more effective than analyzing single words. The best performing models used word2vec \cite{mikolov_efficient_2013} which make use of word embeddings to represent sequences of words and can be used to analyse text. However, word2vec lacks the ability to analyze full sentences or longer range dependencies.


\begin{wrapfigure}{r}{0.5\textwidth}
  \includegraphics[width=0.5\textwidth]{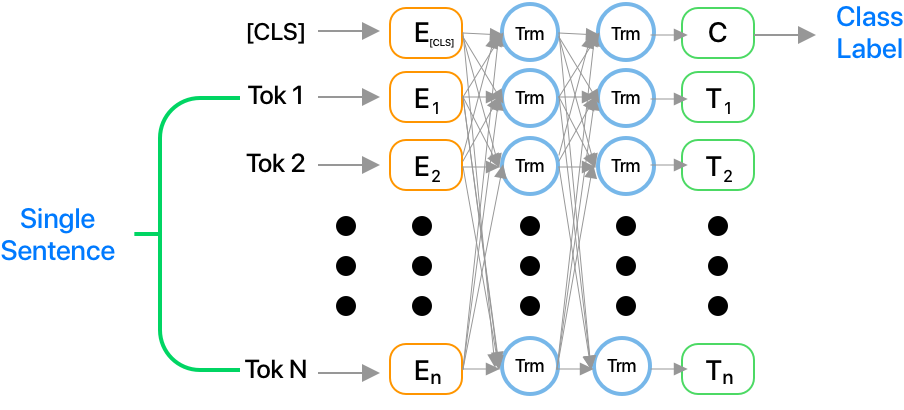}
  \caption[width=0.5\textwidth]{BERT architecture for sentence classification task}
  \label{fig:bert-sentence}
\end{wrapfigure}

Current NLP research is being dominated by the use of bidirectional transformer models such as BERT \cite{devlin_bert_2019}. Transformer models use word embeddings as input similar to word2vec; however the models can handle longer input sequences and the relations within these sequences. This ability, combined with the attention mechanism described in the famous \textit{"attention is all you need"} paper \cite{vaswani2017attention} enables BERT to find long range dependencies in text leading to more robust language models. 

All top 10 submissions for the GLUE benchmark \cite{wang2018glue} make use of BERT models, thus it would be intuitive to conclude it would be interesting to use a BERT model as text analysis model for our task. Figure \ref{fig:bert-sentence} shows a BERT architecture for sentence classification.

The original BERT model was pre-trained on a large quantity of multilingual data. However, since the open sourcing of the BERT architecture by Google, a multitude new models have been made available including monolingual models constructed for tasks in specific languages. \cite{martin2019camembert}\cite{Kuratov2019AdaptationOD}\cite{virtanen2019multilingual}\cite{Antoun2020AraBERTTM}
A comparison of monolingual BERT model performance and multilingual BERT model performance \cite{nozza_what_2020} on various tasks showed that monolingual BERT models outperform multilingual models on every task table \ref{tab:BERT-mono-vs-multi} shows a short summary of their evaluation as performed by Nozza et al.

\begin{table}
  \centering
  \begin{tabular}{@{}lcccr@{}}
  \toprule
    \textbf{Task} & \textbf{Metric} & \textbf{Avg. Monolingual BERT} & \textbf{Avg. Multilingual BERT} & \textbf{Diff}\\ \midrule
    Sentiment Analysis & Accuracy & \textbf{90.17} \% & 83.80 \% & 6.37 \% \\
    Text Classification & Accuracy & \textbf{88.96} \% & 85.22 \% & 3.75 \% \\ \bottomrule
   \end{tabular}
   \caption{Different monolingual BERT model average performance on various tasks versus multilingual BERT  \cite{nozza_what_2020}}
 \label{tab:BERT-mono-vs-multi}
\end{table}

for the Dutch language the top performing models are RobBERT \cite{delobelle_robbert_2020} which is a BERT model using a different set of hyperparameters as described by Yinhan Liu, et al. \cite{liu2019roberta} This model architecture is dubbed RoBERTa. The other model BERTje \cite{de_vries_bertje_2019} is more traditional in the sense that the pretraining hyperparameters follow the parameters as described in the original BERT publication. Table \ref{tab:top-dutch-bert} provides a short overview of these models

\begin{table}[h!]
    \centering
    \begin{tabular}{@{}lccr@{}} \toprule
       \textbf{Model name} & \textbf{Pretrain corpus} & \textbf{Tokenizer type} & \textbf{Acc Sentiment analysis} \\ \midrule
         belabBERT  & Common Crawl Dutch (non-shuffled) & BytePairEncoding & 95.92$^{*}$ \% \\
         RobBERT  & Common Crawl Dutch (shuffled) & BytePairEncoding & 94.42 \% \\
         BERTje  & Mixed (Books, Wikipedia, etc) & Wordpiece & 93.00 \% \\ \bottomrule
    \end{tabular}
    \caption{The 3 top performing monolingual dutch BERT models based on their sentiment analysis accuracy \cite{delobelle_robbert_2020} \\ \textsuperscript{*} to be verified }
    \label{tab:top-dutch-bert}
\end{table}

\section{Audio classification}

As highlighted in the introduction, the field of computational audio analysis is well established. Most researchers extract speech parameters from raw audio and base their classification on this. Speech parameters reflect important brain functions such as motor speed (articulation), emotional status (prosody), cognitive functioning (correct use of grammar, vocabulary scope) and social behavior (timbre matching),

Pause length, and percentage of pauses were found to be highly correlated with psychotic symptoms  \cite{cohen_psychiatric_2013}.
Marmar et al. identified several Mel-frequency cepstral coefficients (MFCC) which are highly indicative for depression \cite{marmar_speech-based_2019}.

The features described in these papers can be quantitatively extracted from speech samples. We assume these features to also be indicative for our classification task as both groups are included.

\chapter{Methods}

As highlighted in the introduction, we aim to create a model that is able to perform classification based on only the text. Later on we show how this could be extended to a hybrid form, for this hybrid model we use a simple audio classification network. In this chapter we present a hybrid model that uses a the BERT based architecture for text classification. We use the top performing Dutch model RobBERT and a novel trained RoBERTa based model called belabBERT. For the audio analysis we use a simple neural network. Finally, we combine the output of these models in the hybrid network

\section{Data preprocessing}

Of the 339 interviews, 141 were transcribed, of which were 76 psychotic, 6 depressive and 59 healthy participants. Transcripts were transformed from the CHAT format to flat text. Since we are dealing with privacy-sensitive information we took measures to mitigate any risk of leaking sensitive info. For audio we only perform analysis on parameters that were derived from the raw audio, not including any content. For the transcripts we swapped all transcripts with their tokenized versions and only performed calculations on these. In order to create more examples, full tokenized transcripts were chunked into a length of 220 tokens per chunk and 505 tokens per chunk resulting in two transcript datasets per tokenizer table \ref{tab:total-samples-transcripts} shows the amount of samples after chunking.

The acquired datasets were split into 80\% training set, 10 \% validation and 10 \% test set keeping the ratios among participants of the original dataset.

\begin{table}[h!]
    \centering
    \begin{tabular}{@{}lccccr@{}}
        \toprule
        \textbf{Dataset ID}  &  \textbf{Chunk size} & \textbf{Psychotic} & \textbf{Control} & \textbf{Depressive} & \textbf{Total} \\
        \midrule
         belabBERT-505 & \cellcolor[HTML]{FFE57C} 505 & 294 & 274 & 24 & 592 \\
         belabBERT-220 & \cellcolor[HTML]{FFDA45} 220 & 625 & 589 & 52 & 1266 \\
         RobBERT-505  &\cellcolor[HTML]{FFE57C} 505 & 499 & 127 & 41 & 1012 \\
         RobBERT-220  &\cellcolor[HTML]{FFDA45} 220 & 1096 & 1043 & 92 & 2231 \\ \midrule
         \textbf{Full} & --- & 76 & 59 & 6 & 141 \\ \bottomrule
    \end{tabular}

    \caption{Total amount of samples after chunking with different chunk lengths and different tokenizers}
    \label{tab:total-samples-transcripts}
\end{table}

\section{Text classification}


\subsection{belabBERT}

We hypothesize that a language model which is pretrained on data that resembles the data of its fine tuning task (text classification of transcripts in our case) will perform better then general models. Our dataset consists interview transcripts thus conversational data. The problem is that RobBERT was pretrained on a shuffled version of the the OSCAR Web crawl corpus. This limits the range over which RobBERT can find relations between words, RobBERT also uses the RoBERTa base tokenizer which is a tokenizer trained on a English corpus, we assumed this affects the performance of RobBERT negatively on downstream tasks. since the previously referenced meta-analysis \cite{de_boer_clinical_2018} recommends future research looks at models which are able to analyze larger group of words, sentences to be specific. 

We decided to train a RoBERTa based Dutch language model from scratch on the non-shuffled OSCAR corpus \cite{ortiz-suarez-etal-2020-monolingual} which consists of a set of monolingual corpora extracted from Common Crawl snapshots. We also trained a byte pair encoding tokenizer on the same corpus to create the word embeddings which belabBERT uses as input, alleviating potential problems in RobBERT both regarding tokenizer as well as long-term dependencies. We use the original RoBERTa training parameters

\subsection{Fine tuning}

\begin{figure}[h!]
\centering
  \includegraphics[width=0.75\textwidth]{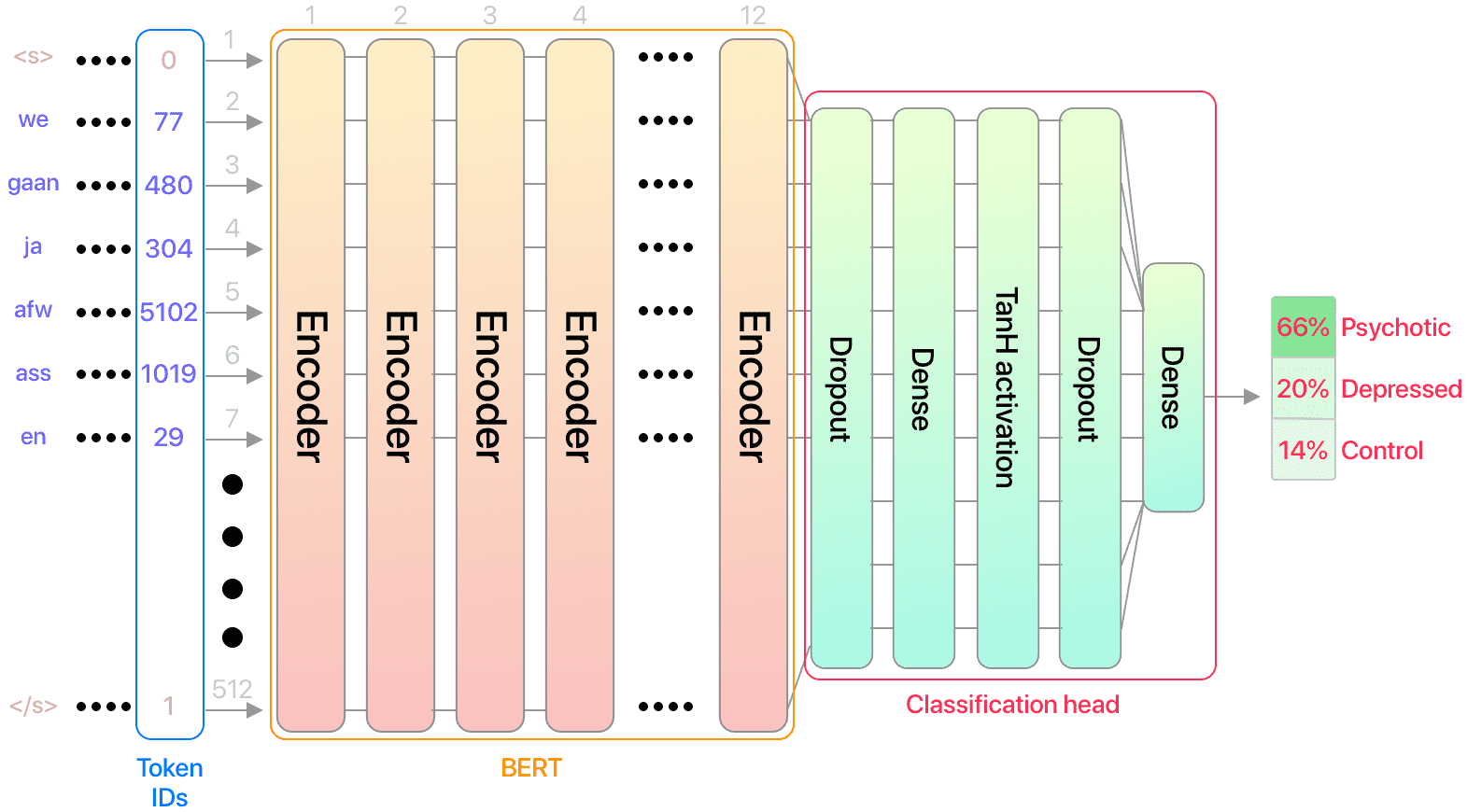}
  \caption[width=0.75\textwidth]{Model architecture for text classification, green marked text are regular tokens and the pink text marks the special tokens indicating a begin of sentence token}
  \label{fig:bert-classification-fine-tuning}
\end{figure}

In order to fine tune belabBERT and RobBERT for the classification of text input we implemented the classifier head as described in the BERT paper a visualization can be found in figure \ref{fig:bert-classification-fine-tuning} the output layer consists of 3 output neurons. In order to find the optimal hyperparameter set we performed several runs with different sets of configurations. In the results chapter we will go more in depth about the specifics of this process.

\newpage
\section{Audio analysis}

\begin{figure}[h!]
\centering
  \includegraphics[width=0.75\textwidth]{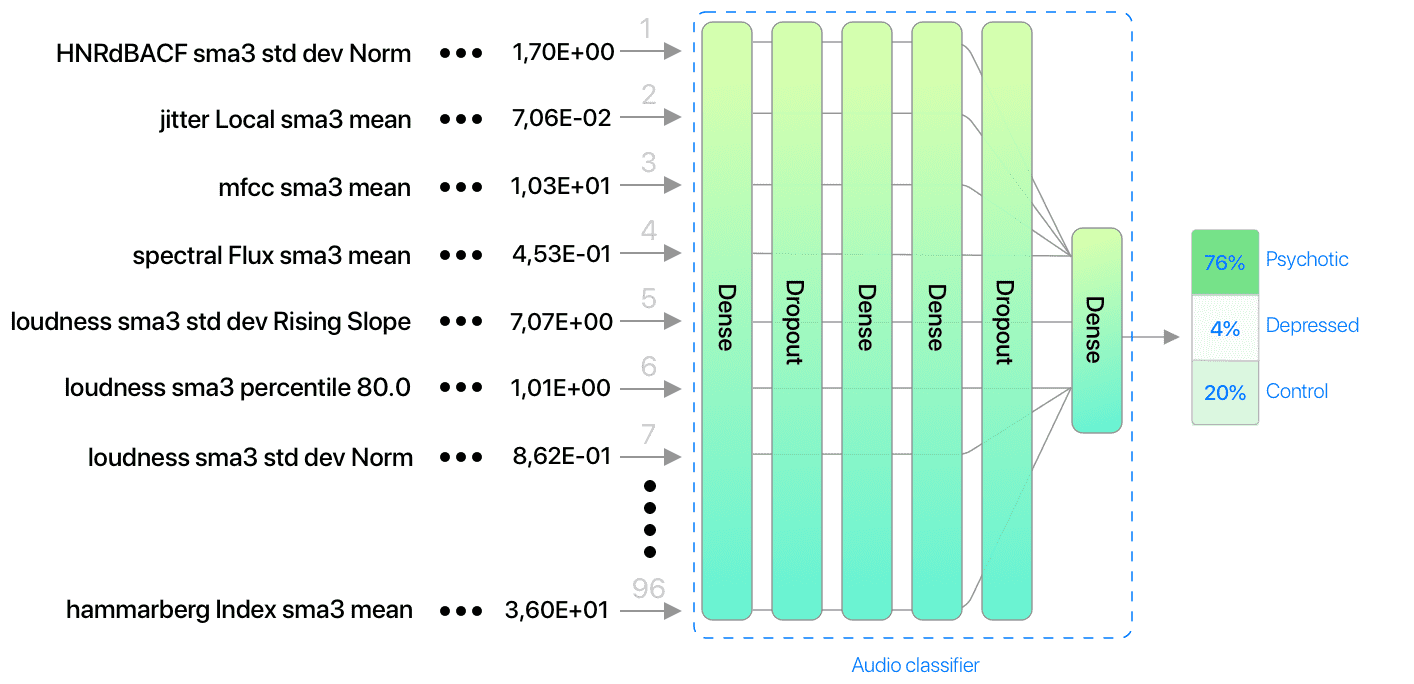}
  \caption[width=0.75\textwidth]{Model architecture for audio classification based on extracted speech features}
  \label{fig:audio-class}
\end{figure}

Related work in audio analysis for diagnostic purposes found that impressive results can be achieved using speech parameters only. Our dataset provides us of a pre-processed set of speech parameters for every audio interview. These are extracted using openSMILE and the eGeMAPS package \cite{eyben_opensmile_2010}. Using this set of features, we use a simple neural network architecture consisting of three layers of which the specifics can be seen in figure \ref{fig:audio-class}. The majority of research in this field focuses on more traditional machine learning techniques such as logistic regression or support vector machine.
However, these are less resistant to noise in the data and thus require feature engineering before processing the parameters. A notable weakness of feature engineering is that information is lost, as it is difficult for traditional machine learning techniques to cope with noise that irrelevant features provide. Using a neural network enables us to use all audio extracted speech parameters as input and automatically learn which features are relevant for each classification

\section{Hybrid model}
\begin{figure}[h!]
\centering
  \includegraphics[width=\textwidth]{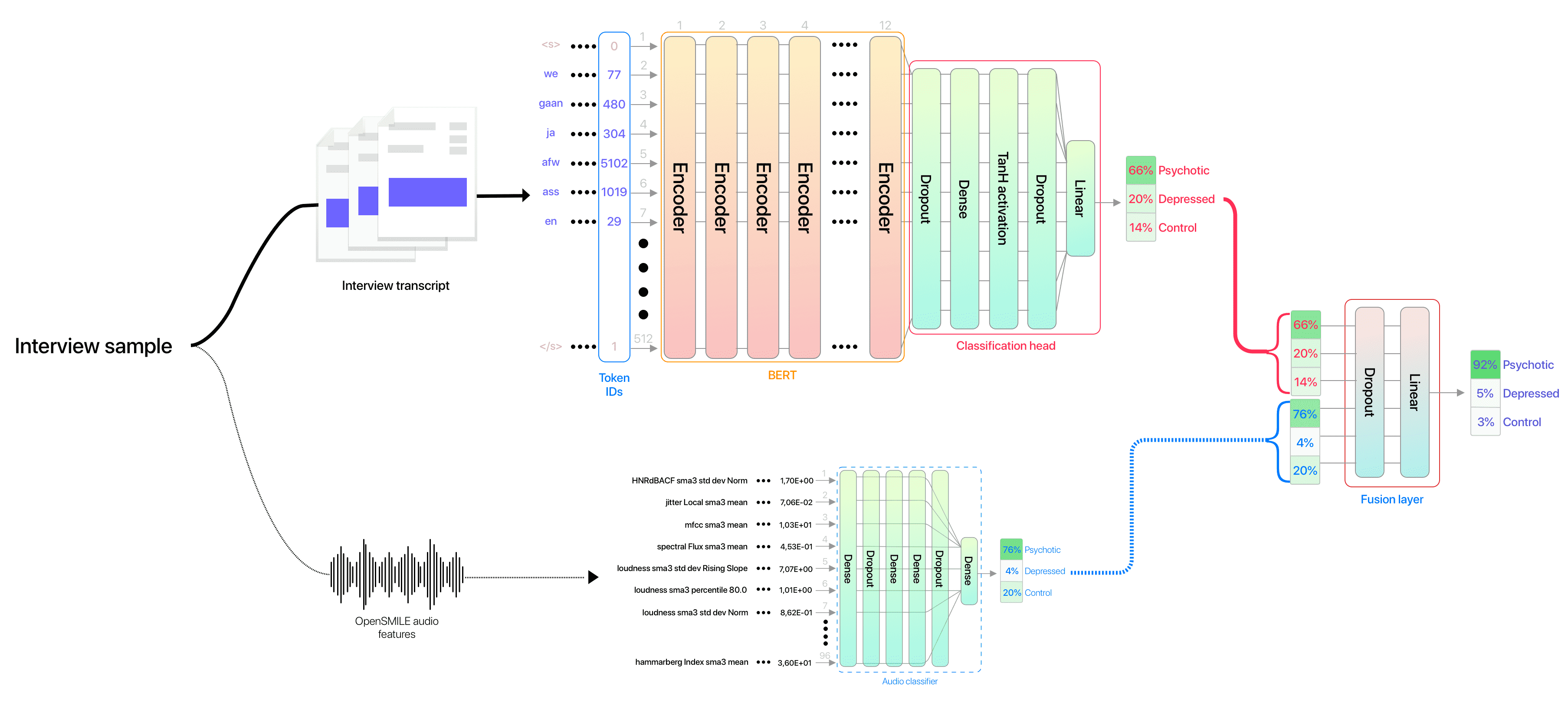}
  \caption[width=0.5\textwidth]{hybrid model architecture for classification task}
  \label{fig:bert-hybrid}
\end{figure}

We developed a hybrid model making use of both modalities (text and audio) and compared its performance to the single models. We assume this model improves the accuracy of the classification since audio characteristics are not embedded in text data; e.g. variations in pitch can be highly indicative for depression \cite{marmar_speech-based_2019} however this is parameter is not present in text data.
Similarly, coherence of grammar and semantic dependencies are indicative of the mental state of a person but is not found in the audio signal. There are multiple ways and techniques to combine models. As this thesis aims to present an initial proof of concept for hybridisation we stick to a simple "late fusion" architecture with a fully-connected layer to map the output of both models into 3 outputs. After training both models separately weights will be frozen and output layers of the separate models will be used to generate inputs for the hybrid model. Figure \ref{fig:bert-hybrid} shows an overview of this combined model.

\chapter{Experiments}

This chapter shows the results of our experiments. In the text analysis section we compare the performances of the proposed belabBERT against RobBERT, the best performing model will be used as input for our fusion model. 

\section{Experimental setup}
All experiments were run on a high performance computing cluster. The language model belabBERT was trained on 16 Nvidia Titan RTX GPUs (24GB each) for a total of 60 hours. All other tasks were run on a single node containing 4 GPUs of the same specifications. 

\subsection{Pretraining corpus}
For the pretraining of belabBERT we used the OSCAR corpus  \cite{ortiz-suarez-etal-2020-monolingual} which consists of a set of monolingual corpora extracted from Common Crawl snapshots. For this thesis a non-shuffled version was made available for the Dutch corpus, which consists of 41GB raw text. This is in contrast with the corpus used for RobBERT, which uses the shuffled and pre-cleaned version. By using a non-shuffled version the sentence order of the corpus is preserved. This property hopefully enables belabBERT to learn long range syntactic dependencies. On top of that, we perform a sequence of common preprocessing steps in order to better match the source of our interview transcript data. These preprocessing steps included, fuzzy deduplication (i.e remove lines with a +90\% overlap with other lines), removing non textual data such as \textit{"https://"} and excluding lines longer than 2000 words. this resulted in a total amount of 32GB clean text of which 10\% was held-out as validation set to accurately measure overfitting.

\subsection{Implementation}

\subsubsection{belabBERT}

The language model belabBERT was created using the Hugging Face's transformer library\cite{Wolf2019HuggingFacesTS}, a Python library which provides a lot of boilerplate code for building BERT models. belabBERT uses a RoBERTa architecture \cite{liu2019roberta}, unless otherwise specified all parameters for the training of this model are kept default. The model and used code is publicly available under an MIT open-source license on \href{https://github.com/Joppewouts/belabBERT}{GitHub}

\subsubsection{Remaining models}
All other models used in this thesis (text classifier, audio classifier and hybrid classifier) are developed in Python using the PyTorch Lightning \cite{falcon2019pytorch} framework. Hyperparameter optimization was performed using the Weights \& Biases Sweeps system \cite{wandb}. This process involves generating a large set of configuration parameters based on pre-defined default parameter values and training the model accordingly, we picked the model with the lowest cross-entropy loss on the held-out validation set assuming this model is best generalisable.

\section{Training configurations}

The core experiments for this thesis are based on the configurations of subsections \ref{subsec:belabBERT} and \ref{subsec:RobBERT}.
To measure the effect of chunk sizes we ran two separate analyses for each base model (belabBERT and RobBERT), with a varying chunk size of 220 and 505 tested for each model. A dutch BPE tokenizer is used for belabBERT to create its word embeddings which makes it an efficient tokenizer for our dataset when compared to the Multi lingual tokenizer used for RoBERTa. As a consequence, belabBERT produces less tokens for a Dutch text than RobBERT which explains the skewed sizes of training samples. Our default hyperparameters follow the GLUE fine tuning parameters used in the original RoBERTa paper \cite{liu2019roberta}. 
Subsection \ref{subsec:extend-to-hybrid} shows the training configuration which was used for the hybrid model, this involves two neural networks which were trained separately, in which the first described model takes audio features as input, the second is the fusion layer which bases its output classification on 6 tensorized input values. In order to find the optimal set of hyperparameters we train each model 15 times. We show the parameter set for the described model that reached the lowest cross-entropy validation loss. The results are presented in chapter \ref{chap:five}.

\subsection{belabBERT} \label{subsec:belabBERT}

We train belabBERT in the two different chunk sizes, 505 and 220. We expect belabBERT to outperform RobBERT due to the nature of its pretraining corpus and custom Dutch tokenizer.

\subsubsection{chunk size 505}

\begin{table}[h!]
\centering
\begin{tabular}{@{}lcccr@{}}
\toprule
\textbf{Set}                    & \textbf{Psychotic}       & \textbf{Depressed}      & \textbf{Healthy} & \textbf{\% Of total}\\ \midrule
Train      & 235 & 19 & 219 & 80\%           \\
Validation & 29  & 2  & 27  & 10\%         \\
Test       & 30  & 3  & 28  & 10\%         \\ \midrule
Total      & 294 & 24 & 274 & 100\%        \\ \bottomrule
\end{tabular}
\caption{\textbf{Overview of samples per category for training belabBERT with 505 chunk size}}
\label{tab:belabBERT-505-train-val-test}
\end{table}

\begin{table}[h!]
    \centering
    \begin{tabular}{@{}l|c@{}}
\toprule
\textbf{Parameter name} & \textbf{Value} \\ \midrule
Batch size              & 10             \\
Epochs                  & 3              \\
Peak learning rate      & $6.22e^{-5}$   \\
Warmup steps            & 373            \\ \bottomrule
\end{tabular}
    \caption{\textbf{Parameters for best performing model belabBERT with 505 chunk size}}
    \label{tab:belabBERT-505-parameters}
\end{table}

\subsubsection{chunk size 220}

\begin{table}[h!]
    \centering
  \begin{tabular}{@{}lcccr@{}} 
\toprule
       \textbf{Set} & \textbf{Psychotic} & \textbf{Control} & \textbf{Depressed} & \textbf{\% Of total} \\ \midrule
         Train & 500 & 471 & 41 & 80\% \\
         Validation & 62 & 59 & 5 & 10\% \\
         Test & 63 & 59 & 6 & 10\% \\ \midrule
         Total & 625 & 589 & 52 & 100\% \\ \bottomrule
    \end{tabular}
    \caption{\textbf{Overview of samples per category for training belabBERT with 220 chunk size}}
    \label{tab:belabBERT-220-train-val-test}
\end{table}

\begin{table}[h!]
    \centering
        \begin{tabular}{@{}l|c@{}}
        \toprule
       \textbf{Parameter name} & \textbf{Value} \\ \midrule
         Batch size & 9 \\
         Epochs & 5 \\
         Peak learning rate & $8.42e^{-5}$ \\
         Warmup steps & 190 \\ \bottomrule
    \end{tabular}
    \caption{\textbf{Parameters for best performing model belabBERT with 220 chunk size}}
    \label{tab:belabBERT-220-parameters}
\end{table}
\newpage

\subsection{RobBERT} \label{subsec:RobBERT}

In order to evaluate the performance of belabBERT we evaluate it against the performance of the current Dutch state-of-the-art model RobBERT. The results of these experiments will help us to better contextualize the achieved results of belabBERT.

\subsubsection{chunk size 505}

\begin{table}[h!]
    \centering
    \begin{tabular}{@{}lcccr@{}} 
\toprule
       \textbf{Set} & \textbf{Psychotic} & \textbf{Control} & \textbf{Depressed} & \textbf{\% of total} \\ \midrule
         Train & 398 & 100 & 31 & 80\% \\
         Validation & 50 & 13 & 5 & 10\% \\
         Test & 51 & 14 & 5 & 10\% \\ \midrule
         Total & 499 & 127 & 41 & 100\% \\ \bottomrule
    \end{tabular}
    \caption{\textbf{Overview of samples per category for training RobBERT with 505 chunk size}}
    \label{tab:robbert-505-train-val-test}
\end{table}

\begin{table}[h!]
    \centering
    \begin{tabular}{@{}l|c@{}}
        \toprule
       \textbf{Parameter name} & \textbf{Value} \\ \midrule
         Batch size & 10\% \\
         Epochs & 3 \\
         Peak learning rate & $1.19e^{-4}$ \\
         Warmup steps & 401 \\ \bottomrule
    \end{tabular}
    \caption{\textbf{Parameters for best performing model RobBERT with 505 chunk size}}
    \label{tab:robbert-505-parameters}
\end{table}

\subsubsection{chunk size 220}

\begin{table}[h!]
    \centering
        \begin{tabular}{@{}lcccr@{}} 
        \toprule
       \textbf{Set} & \textbf{Psychotic} & \textbf{Control} & \textbf{Depressed} & \textbf{\% Of total}\\ \midrule
         Train & 876 & 834 & 73 & 80\% \\
         Validation & 110 & 104 & 9 & 10\% \\
         Test & 110 & 105 & 10 & 10\% \\ \midrule
         Total & 1096 & 1043 & 92 & 100\% \\ \bottomrule
    \end{tabular}
    \caption{\textbf{Overview of samples per category for training RobBERT with 220 chunk size.}}
    \label{tab:robbert-220-train-val-test}
\end{table}

\begin{table}[h!]
    \centering
    \begin{tabular}{@{}l|c@{}}
        \toprule
       \textbf{Parameter name} & \textbf{Value} \\ \midrule
         Batch size & 13 \\
         Epochs & 3 \\
         Peak learning rate & $6.58e^{-5}$ \\
         Warmup steps & 401 \\ \bottomrule
    \end{tabular}
    \caption{\textbf{Parameters for the best performing RobBERT model with 220 chunk size}}
    \label{tab:robbert-220-parameters}
\end{table}

\subsection{Extending to a hybrid model} \label{subsec:extend-to-hybrid}

The hybrid model consists of a separately trained audio classification network.
In order to maximize the size of available training samples for the fusion we trained the audio classifier on samples of which no transcript was available. The held-out test set of our audio classifier consists of all samples of which a transcript did exist, this makes sure there is no overlap between the training data of the audio classifier and the text classifier.

\subsubsection{Audio classification}

The audio classification network uses categorical cross-entropy loss and Adam optimization\cite{kingma2014adam} with $\beta_{1} = 0.9, \beta_{2} = 0.95$ and $\epsilon = 10^{-8}$, due to the inherent noisy nature of an audio signal and its extracted features we use a default dropout rate of $0.1$. The learning rate boundaries were found by performing a initial training run in during which, the learning rate linearly increases for each epoch as described by L. Smith \cite{smith2017cyclical}. We picked the median learning rate of these bounds as our default learning rate

\begin{table}[h!]
    \centering
    \begin{tabular}{@{}lcccc@{}}
\toprule
       \textbf{Set} & \textbf{Psychotic} & \textbf{Control} & \textbf{Depressed} & \textbf{\% Of total} \\ \midrule
         Train & 97 & 74 & 7 & 53 \\
         Validation & 10 & 8 & 2 & 6 \\
         Test & 76 & 59 & 6 & 41 \\ \midrule
         Total & 183 & 141 & 15 & 100\% \\ \bottomrule
    \end{tabular}
    \caption{Overview of samples per category for training Audio classification network}
    \label{tab:audio-train-test}
\end{table}

\begin{table}[h!]
    \centering
    \begin{tabular}{@{}l|c|c@{}}
        \toprule
       \textbf{Parameter name} & \textbf{Default} & \textbf{Best} \\ \midrule
         Batch size & 4 & 15 \\
         Epochs & 10 & 50 \\
         Learning rate & $2.5e^{-2}$ & $5e^{-7}$ \\
         Dropout rate & 0.1 & 0.3 \\ \bottomrule
    \end{tabular}
    \caption{\textbf{Default and best performing parameters for the audio classification network}}
    \label{tab:audio-classification-parameters}
\end{table}

\subsubsection{Hybrid classification}

We trained the hybrid classification on the dataset of our best performing text classification network, its important to remember that due to the chunking of this dataset we have multiple samples stemming from a single patient which is discussed in chapter \ref{chap:five}, this explains the difference in total amount of samples between the audio classification and hybrid classification. The train/validate/test dataset used for the hybrid classifier is shown in Table \ref{tab:belabBERT-220-train-val-test}

\begin{table}[h!]
    \centering
    \begin{tabular}{@{}l|c@{}}
        \toprule
       \textbf{Parameter name} & \textbf{Value} \\ \midrule
         Batch size & 16 \\
         Epochs & 55 \\
         Learning rate & $1e^{-2}$ \\
         Dropout rate & 0.15 \\ \bottomrule
    \end{tabular}
    \caption{Parameters for best performing hybrid classification network}
    \label{tab:hybrid-classification}
\end{table}

\chapter{Results} \label{chap:five}
In this chapter we present the results for the previously described experiments. After each section we evaluate the results, in the last section of this chapter we discuss the overall results

\section{belabBERT and RobBERT} \label{sec:blabenrob}
This section presents the results of subsection \ref{subsec:belabBERT} and \ref{subsec:RobBERT}, for the overall best performing model we show additional common classification metrics.

\subsection{Results}

Table \ref{tab:Results-text-classification} shows that both experiments with belabBERT as its base model manages to outperform the current Dutch state-of-the-art RobBERT with the top performing model using a chunk size of 220 achieving a classification accuracy of 75.68\% on the test set and 71.18\% on validation set. The top performing model with RobBERT as base also uses a chunk size of 220 and reaches a 69.06\% classification accuracy on the test set and 69.64\% on the validation set.

\begin{table}[h!]
    \centering
    \begin{tabular}{@{}lcc@{}}
\toprule
       \textbf{Experiment} & \textbf{Validation accuracy} & \textbf{Test accuracy} \\ \midrule
         belabBERT 505 & 70.25\% & 73.91\% \\
         belabBERT 220 & \textbf{71.18\%} & \textbf{75.68\%} \\
         RobBERT 505  & 68.93\% & 65.69\% \\
         RobBERT 220  & 69.64\% & 69.06\% \\ \bottomrule
    \end{tabular}
    \caption{Classification accuracy for the best performing belabBERT and RobBERT based models on the held-out validation and test set}
    \label{tab:Results-text-classification}
\end{table}

\begin{figure}[h!]
    \centering
  \includegraphics[width=0.5\textwidth]{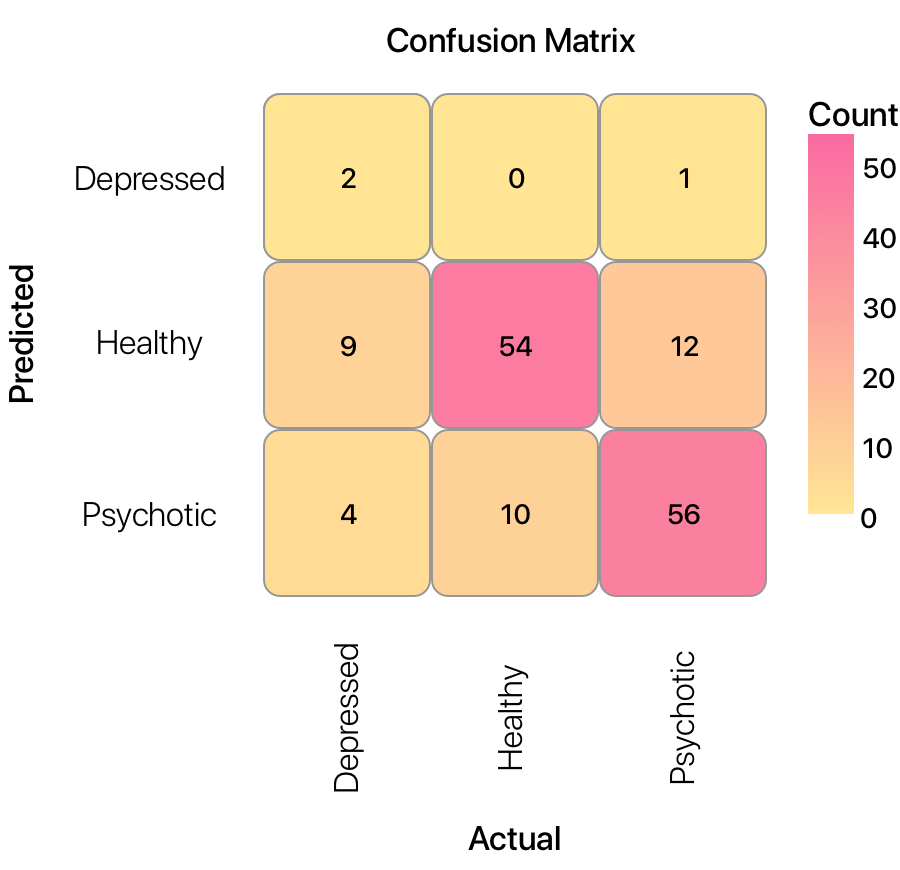}
  \caption[width=0.5\textwidth]{Results for the belabBERT based model with a chunk size of 220:  Predicted classes vs Actual classes}
  \label{fig:confuse_text}
\end{figure}

\begin{table}[h!]
    \centering
    \begin{tabular}{@{}lccc@{}}
    \toprule
       \textbf{Metric} & \textbf{Depressed} & \textbf{Healthy} & \textbf{Psychotic} \\ \midrule
         Recall & 13.33\% & \textbf{84.38\%} & 81.16\% \\
         Precision & 66.67\% & 72.00\% & \textbf{80.00\%} \\ 
         F1-score & 22.21\% & 77.70\% & \textbf{80.58\%} \\ \bottomrule
    \end{tabular}
    \caption{Classification metrics for the belabBERT based model with a chunk size of 220}
    \label{tab:metrics-text-classification}
\end{table}

\newpage
\subsection{Evaluation}

The results shown in \ref{tab:Results-text-classification} confirm our initial hypothesis, belabBERT does indeed benefit from its ability to capture long range semantic dependencies. Both on the 505 chunk size, as well as the 220 chunk size experiments belabBERT manages to outperform the current state-of-the-art language model RobBERT. belabBERT 220 has a limited recall for the depression label but its precision is higher than expected.
\newpage
\section{Extending to a hybrid model} \label{sec:ext-to-hybrid}
In this section we present the audio classification results and the results which is part of the extension towards the hybrid classification network which uses the best performing text classification network.
\subsection{Results}

\subsubsection{Audio classification}
\begin{figure}[h!]
\centering
  \includegraphics[width=0.5\textwidth]{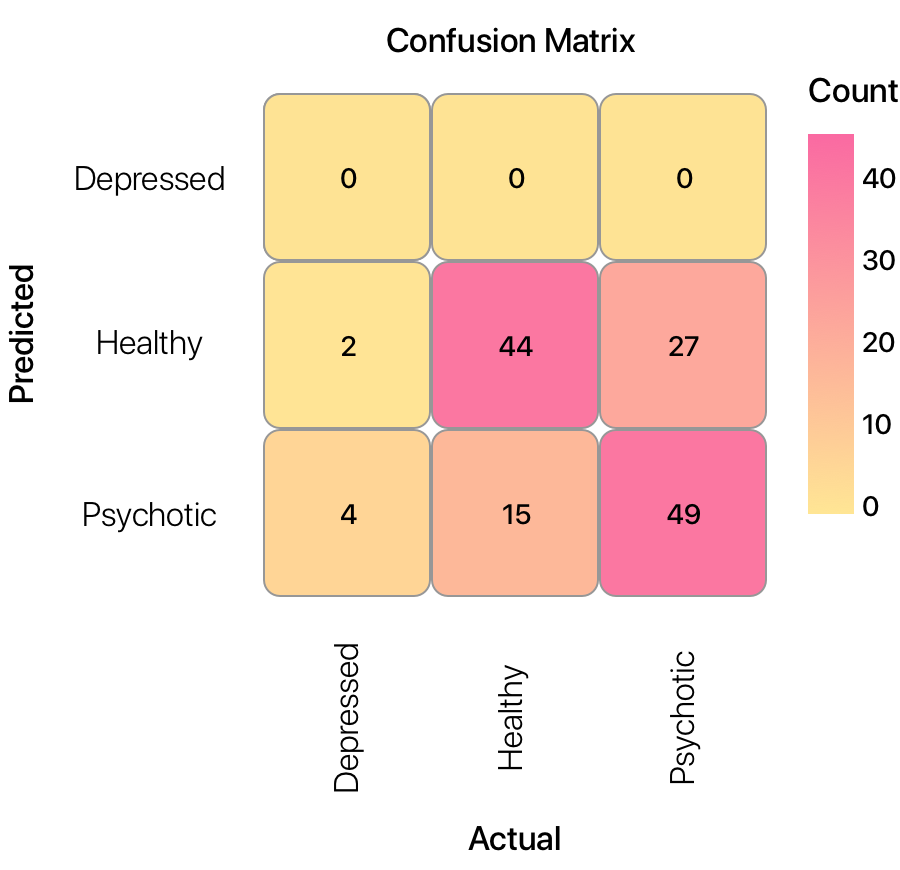}
  \caption[width=0.5\textwidth]{Audio classifier results:  Predicted classes vs Actual classes}
  \label{fig:confuse_audio}
\end{figure}

Table \ref{tab:audio-class-test-val} shows the audio classification network reached a classification accuracy of 65.96 \% on the test set and 80.05\% accuracy on the validation set, due to the small size of this set we should not consider this result as significant, we also observe in \ref{fig:confuse_audio} that the network was not able to distinguish samples with the depressed label from the other labels based on its inputs.

\begin{table}[h!]
    \centering
    \begin{tabular}{@{}ccc@{}}
\toprule
        \textbf{Validation accuracy} & \textbf{Test accuracy} \\ \midrule
        80.05\textsuperscript{*}\% & 65.96\% \\ \bottomrule
    \end{tabular}
    \caption{Classification accuracy of the audio classification network on the held-out validation and test set \\ \textsuperscript{*} validation set size was very small}
    \label{tab:audio-class-test-val}
\end{table}

\begin{table}[h!]
    \centering
    \begin{tabular}{@{}lccc@{}}
    \toprule
       \textbf{Metric} & \textbf{Depressed} & \textbf{Healthy} & \textbf{Psychotic} \\ \midrule
         Recall & 0\% & \textbf{75.6\%} & 64.47\% \\
         Precision & 0\% & 60.27\% &\textbf{ 72.05\%} \\ 
         F1-score & 0\% & 67.07\% & \textbf{68.05\%} \\ \bottomrule
    \end{tabular}
    \caption{Classification metrics for audio classification}
    \label{tab:metrics-audio-classification}
\end{table}

\subsubsection{Hybrid classification} \label{subsubsec:hybridcla}

Table \ref{tab:hybrid-class-test-val} shows the classification accuracies for the hybrid classification network, it reaches an accuracy of 77.70\% on the test set and a 70.47\% accuracy on the validation set. 

\begin{table}[h!]
    \centering
    \begin{tabular}{@{}ccc@{}}
\toprule
        \textbf{Validation accuracy} & \textbf{Test accuracy} \\ \midrule
        70.47\% & 77.70\% \\ \bottomrule
    \end{tabular}
    \caption{Classification accuracy of the hybrid classification network on the held-out validation and test set}
    \label{tab:hybrid-class-test-val}
\end{table}

\begin{figure}[h!]
\centering
  \includegraphics[width=0.5\textwidth]{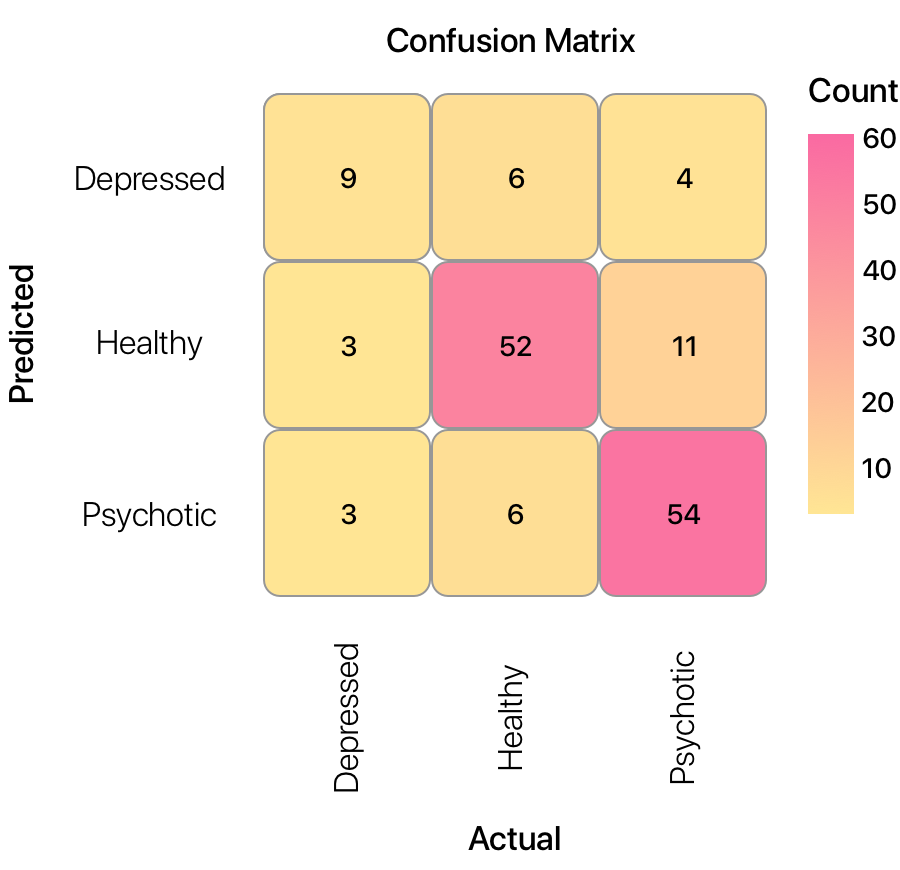}
  \caption[width=0.5\textwidth]{Hybrid classifier results: Predicted classes vs Actual classes}
  \label{fig:fusion_confuse}
\end{figure}

\begin{table}[h!]
    \centering
    \begin{tabular}{@{}lccc@{}}
    \toprule
       \textbf{Metric} & \textbf{Depressed} & \textbf{Healthy} & \textbf{Psychotic} \\ \midrule
         Recall & 60.00\% & \textbf{81.25\%} & 78.26\% \\
         Precision & 47.37\% & 78.79\% & \textbf{85.71\%} \\ 
         F1-score & 52.94\% & 80.01\% & \textbf{81.82\%} \\ \bottomrule
    \end{tabular}
    \caption{Classification metrics for hybrid classification}
    \label{tab:metrics-fusion-classification}
\end{table}

\subsection{Evaluation}
From our observations of the audio classification network we can conclude that it does not perform that well for the classification of all labels, it does however perform relatively well on the healthy category. The extension towards the hybrid model where we base our classification on both text and audio does however result in an improved classification accuracy. 

\section{Discussion}
From the results in table \ref{tab:Results-text-classification} we can conclude that our self trained model belabBERT reaches a $6.62\%$ higher classification accuracy on the test-set than the best performing RobBERT model. Furthermore, we observe that a smaller chunk size of 220 tokens leads to a significant accuracy gain for both base models. The small difference between the validation and test set accuracies shown in table \ref{tab:Results-text-classification} are a positive indicator that the classification accuracy is significant and representative for the capability of the model to categorize the given text samples. From the difference in classification accuracy between belabBERT and RobBERT we conclude that a BERT model using a specialized Dutch tokenizer and pretrain corpus which resembles on conversational data provides significant benefits on downstream classification tasks. On top of that, we conclude that using a smaller chunk size has a positive effect on the classification accuracy. 

Our brief exploration into the hybridisation of belabBERT with a very basic audio-classification network has pushed its test set accuracy of 75.68\% to a 77.0\% accuracy. From our observations of the classification metrics shown in table \ref{tab:metrics-fusion-classification} we showed that the addition of an audio classification network next to the strong stand-alone text classification model leads to an overall better precision for all labels on top of the higher classification accuracy.
However, the lack of 'depressed' samples in our dataset hinders us from making definitive conclusions about relevance of our findings in this category.

\chapter{Conclusion \& Future work}

\section{Conclusion}
In this thesis, we presented a strong text classification model which challenges the current state of the art audio classification networks used for the classification of psychiatric illness. We introduced a new model belabBERT and showed that this language model which is trained on capturing long range semantic dependencies over sentences in a Dutch corpus outperforms the current state-of-the-art RobBERT model as seen in table \ref{tab:Results-text-classification}. We hypothesized that we could increase the size of our dataset by splitting the samples up into chunks of a fixed length without losing classification accuracy, our results in table \ref{tab:Results-text-classification} support this approach. On top of that we explored the possibilities for a hybrid network which uses both text and audio data as input for the classification of patients as psychotic, depressed or "healthy". Our results in section \ref{subsubsec:hybridcla} indicate this approach is able to improve the accuracy and precision of a stand alone text classification network. Based on these observations we can confirm our main hypothesis that a well designed text-based approach poses a strong competition against the state-of-the- art audio based approaches for the classification of psychiatric illness

\section{Future work}

This section discusses future work on enhancing belabBERT, enhancing the text-based classification of psychiatric illness, possible extensions for the proposed hybrid framework, interpretation and rationalisation of the text classification network.

Compared to BERT models of the same size as belabBERT, it seems that belabBERT is actually still under-trained, the version used during this thesis has only seen 60\% of the training data. Training belabBERT even more could possibly increase its performance on all tasks. 

In our text classification we already applied a chunking technique in order to generate more examples from a single interview sample. However, we observed that prediction accuracy increased when we decreased the chunk size. This leads to the question to explore how the use of even smaller chunk sizes affect the prediction accuracy. When smaller chunk sizes can be used, the amount of training examples is increased, making the model more robust.

While the explored hybrid model we present in this thesis uses pre-extracted audio parameters as input for a neural network it would be interesting to apply new audio analysis techniques. It would be interesting to use raw audio as input for a neural network. The approach would be similar to speech recognition architectures \cite{xiong2016achieving}; a major advantage would be that these architectures can find patterns over time, which makes it possible to discover new relations between input features. 
The hybrid model could also use other data sources to generate a classification such as video which would possibly increase classification accuracy even more

The interpretation and rationalisation of the predictions of neural networks is key for providing clinical relevancy not only in the practical domain of psychiatry but also for the theoretic understanding of the disorder and symptoms. Transformer models like BERT are easily visualisable \cite{coenen2019visualizing}, an extensive interpretation toolkit could provide researchers better tools to discover new patterns in language that are highly indicative for a certain classification prediction, in turn leading to greater understanding of the disorders.

\chapter{Appendix}

\begin{table}[]
    \centering
\begin{longtable}{@{}l@{}}
\toprule
\textbf{Audio parameters}                           \\ \midrule
F0semitoneFrom27.5Hz\_sma3nz\_amean\_numeric        \\
F0semitoneFrom27.5Hz\_sma3nz\_stddevNorm\_numeric   \\
F0semitoneFrom27.5Hz\_sma3nz\_pctlrange0-2\_numeric \\
loudness\_sma3\_amean\_numeric                      \\
loudness\_sma3\_stddevNorm\_numeric                 \\
loudness\_sma3\_pctlrange0-2\_numeric               \\
loudness\_sma3\_meanRisingSlope\_numeric            \\
loudness\_sma3\_stddevRisingSlope\_numeric          \\
loudness\_sma3\_meanFallingSlope\_numeric           \\
loudness\_sma3\_stddevFallingSlope\_numeric         \\
spectralFlux\_sma3\_amean\_numeric                  \\
spectralFlux\_sma3\_stddevNorm\_numeric             \\
mfcc1\_sma3\_amean\_numeric                         \\
mfcc1\_sma3\_stddevNorm\_numeric                    \\
mfcc2\_sma3\_amean\_numeric                         \\
mfcc2\_sma3\_stddevNorm\_numeric                    \\
mfcc3\_sma3\_amean\_numeric                         \\
mfcc3\_sma3\_stddevNorm\_numeric                    \\
mfcc4\_sma3\_amean\_numeric                         \\
mfcc4\_sma3\_stddevNorm\_numeric                    \\
jitterLocal\_sma3nz\_amean\_numeric                 \\
jitterLocal\_sma3nz\_stddevNorm\_numeric            \\
shimmerLocaldB\_sma3nz\_amean\_numeric              \\
shimmerLocaldB\_sma3nz\_stddevNorm\_numeric         \\
HNRdBACF\_sma3nz\_amean\_numeric                    \\
HNRdBACF\_sma3nz\_stddevNorm\_numeric               \\
logRelF0-H1-H2\_sma3nz\_amean\_numeric              \\
logRelF0-H1-H2\_sma3nz\_stddevNorm\_numeric         \\
logRelF0-H1-A3\_sma3nz\_amean\_numeric              \\
logRelF0-H1-A3\_sma3nz\_stddevNorm\_numeric         \\
F1frequency\_sma3nz\_amean\_numeric                 \\
F1frequency\_sma3nz\_stddevNorm\_numeric            \\
F1bandwidth\_sma3nz\_amean\_numeric                 \\
F1bandwidth\_sma3nz\_stddevNorm\_numeric            \\
F1amplitudeLogRelF0\_sma3nz\_amean\_numeric         \\
F1amplitudeLogRelF0\_sma3nz\_stddevNorm\_numeric    \\
F2frequency\_sma3nz\_amean\_numeric                 \\
F2frequency\_sma3nz\_stddevNorm\_numeric            \\
F2amplitudeLogRelF0\_sma3nz\_amean\_numeric         \\
F2amplitudeLogRelF0\_sma3nz\_stddevNorm\_numeric    \\
F3frequency\_sma3nz\_amean\_numeric                 \\
F3frequency\_sma3nz\_stddevNorm\_numeric            \\
F3bandwidth\_sma3nz\_amean\_numeric                 \\
F3bandwidth\_sma3nz\_stddevNorm\_numeric            \\
F3amplitudeLogRelF0\_sma3nz\_amean\_numeric         \\
F3amplitudeLogRelF0\_sma3nz\_stddevNorm\_numeric    \\
alphaRatioV\_sma3nz\_amean\_numeric                 \\
alphaRatioV\_sma3nz\_stddevNorm\_numeric            \\
hammarbergIndexV\_sma3nz\_amean\_numeric            \\
hammarbergIndexV\_sma3nz\_stddevNorm\_numeric       \\
slopeV0-500\_sma3nz\_amean\_numeric                 \\
slopeV500-1500\_sma3nz\_amean\_numeric              \\
slopeV500-1500\_sma3nz\_stddevNorm\_numeric         \\
spectralFluxV\_sma3nz\_amean\_numeric               \\
mfcc1V\_sma3nz\_amean\_numeric                      \\
mfcc1V\_sma3nz\_stddevNorm\_numeric                 \\
mfcc2V\_sma3nz\_amean\_numeric                      \\
mfcc2V\_sma3nz\_stddevNorm\_numeric                 \\
mfcc3V\_sma3nz\_amean\_numeric                      \\
\end{longtable}
    \caption{OpenSMILE Audio features used as input for audio classification network}
    \label{tab:parameters}
\end{table}

\appendix

\bibliography{report}

\begin{thebibliography}{32}
\providecommand{\natexlab}[1]{#1}
\providecommand{\url}[1]{\texttt{#1}}
\expandafter\ifx\csname urlstyle\endcsname\relax
  \providecommand{\doi}[1]{doi: #1}\else
  \providecommand{\doi}{doi: \begingroup \urlstyle{rm}\Url}\fi

\bibitem[Antoun et~al.(2020)Antoun, Baly, and Hajj]{Antoun2020AraBERTTM}
Wissam Antoun, Fady Baly, and Hazem~M. Hajj.
\newblock Arabert: Transformer-based model for arabic language understanding.
\newblock \emph{ArXiv}, abs/2003.00104, 2020.

\bibitem[Biewald(2020)]{wandb}
Lukas Biewald.
\newblock Experiment tracking with weights and biases, 2020.
\newblock URL \url{https://www.wandb.com/}.
\newblock Software available from wandb.com.

\bibitem[Coenen et~al.(2019)Coenen, Reif, Yuan, Kim, Pearce, Vi{\'e}gas, and
  Wattenberg]{coenen2019visualizing}
Andy Coenen, Emily Reif, Ann Yuan, Been Kim, Adam Pearce, Fernanda Vi{\'e}gas,
  and Martin Wattenberg.
\newblock Visualizing and measuring the geometry of bert.
\newblock \emph{arXiv preprint arXiv:1906.02715}, 2019.

\bibitem[Cohen et~al.(2013)Cohen, Kim, and Najolia]{cohen_psychiatric_2013}
Alex~S. Cohen, Yunjung Kim, and Gina~M. Najolia.
\newblock Psychiatric symptom versus neurocognitive correlates of diminished
  expressivity in schizophrenia and mood disorders.
\newblock \emph{Schizophrenia Research}, 146\penalty0 (1):\penalty0 249--253,
  May 2013.
\newblock ISSN 0920-9964.
\newblock \doi{10.1016/j.schres.2013.02.002}.
\newblock URL
  \url{http://www.sciencedirect.com/science/article/pii/S0920996413000777}.

\bibitem[de~Boer et~al.(2018)de~Boer, Voppel, Begemann, Schnack, Wijnen, and
  Sommer]{de_boer_clinical_2018}
J.~N. de~Boer, A.~E. Voppel, M.~J.~H. Begemann, H.~G. Schnack, F.~Wijnen, and
  I.~E.~C. Sommer.
\newblock Clinical use of semantic space models in psychiatry and neurology:
  {A} systematic review and meta-analysis.
\newblock \emph{Neuroscience \& Biobehavioral Reviews}, 93:\penalty0 85--92,
  October 2018.
\newblock ISSN 0149-7634.
\newblock \doi{10.1016/j.neubiorev.2018.06.008}.
\newblock URL
  \url{http://www.sciencedirect.com/science/article/pii/S0149763418301878}.

\bibitem[de~Boer et~al.(2020)de~Boer, Brederoo, Voppel, and
  Sommer]{de_boer_anomalies_2020}
Janna~N. de~Boer, Sanne~G. Brederoo, Alban~E. Voppel, and Iris E.~C. Sommer.
\newblock Anomalies in language as a biomarker for schizophrenia.
\newblock \emph{Current Opinion in Psychiatry}, 33\penalty0 (3):\penalty0
  212--218, May 2020.
\newblock ISSN 1473-6578.
\newblock \doi{10.1097/YCO.0000000000000595}.

\bibitem[de~Vries et~al.(2019)de~Vries, van Cranenburgh, Bisazza, Caselli, van
  Noord, and Nissim]{de_vries_bertje_2019}
Wietse de~Vries, Andreas van Cranenburgh, Arianna Bisazza, Tommaso Caselli,
  Gertjan van Noord, and Malvina Nissim.
\newblock {BERTje}: {A} {Dutch} {BERT} {Model}.
\newblock \emph{arXiv:1912.09582 [cs]}, December 2019.
\newblock URL \url{http://arxiv.org/abs/1912.09582}.
\newblock arXiv: 1912.09582.

\bibitem[Delobelle et~al.(2020)Delobelle, Winters, and
  Berendt]{delobelle_robbert_2020}
Pieter Delobelle, Thomas Winters, and Bettina Berendt.
\newblock {RobBERT}: a {Dutch} {RoBERTa}-based {Language} {Model}.
\newblock \emph{arXiv:2001.06286 [cs]}, January 2020.
\newblock URL \url{http://arxiv.org/abs/2001.06286}.
\newblock arXiv: 2001.06286.

\bibitem[Devlin et~al.(2019)Devlin, Chang, Lee, and
  Toutanova]{devlin_bert_2019}
Jacob Devlin, Ming-Wei Chang, Kenton Lee, and Kristina Toutanova.
\newblock {BERT}: {Pre}-training of {Deep} {Bidirectional} {Transformers} for
  {Language} {Understanding}.
\newblock \emph{arXiv:1810.04805 [cs]}, May 2019.
\newblock URL \url{http://arxiv.org/abs/1810.04805}.
\newblock arXiv: 1810.04805.

\bibitem[Eyben et~al.(2010)Eyben, Wöllmer, and Schuller]{eyben_opensmile_2010}
Florian Eyben, Martin Wöllmer, and Björn Schuller.
\newblock {openSMILE} -- {The} {Munich} {Versatile} and {Fast} {Open}-{Source}
  {Audio} {Feature} {Extractor}.
\newblock pages 1459--1462, January 2010.
\newblock \doi{10.1145/1873951.1874246}.

\bibitem[Eyben et~al.(2015)Eyben, Scherer, Schuller, Sundberg, Andr{\'e},
  Busso, Devillers, Epps, Laukka, Narayanan, et~al.]{eyben2015geneva}
Florian Eyben, Klaus~R Scherer, Bj{\"o}rn~W Schuller, Johan Sundberg, Elisabeth
  Andr{\'e}, Carlos Busso, Laurence~Y Devillers, Julien Epps, Petri Laukka,
  Shrikanth~S Narayanan, et~al.
\newblock The geneva minimalistic acoustic parameter set (gemaps) for voice
  research and affective computing.
\newblock \emph{IEEE transactions on affective computing}, 7\penalty0
  (2):\penalty0 190--202, 2015.

\bibitem[Falcon(2019)]{falcon2019pytorch}
WA~Falcon.
\newblock Pytorch lightning.
\newblock \emph{GitHub. Note:
  https://github.com/PyTorchLightning/pytorch-lightning Cited by}, 3, 2019.

\bibitem[Kingma and Ba(2014)]{kingma2014adam}
Diederik~P Kingma and Jimmy Ba.
\newblock Adam: A method for stochastic optimization.
\newblock \emph{arXiv preprint arXiv:1412.6980}, 2014.

\bibitem[Kuratov and Arkhipov(2019)]{Kuratov2019AdaptationOD}
Yuri Kuratov and Mikhail Arkhipov.
\newblock Adaptation of deep bidirectional multilingual transformers for
  russian language.
\newblock \emph{ArXiv}, abs/1905.07213, 2019.

\bibitem[Liu et~al.(2019)Liu, Ott, Goyal, Du, Joshi, Chen, Levy, Lewis,
  Zettlemoyer, and Stoyanov]{liu2019roberta}
Yinhan Liu, Myle Ott, Naman Goyal, Jingfei Du, Mandar Joshi, Danqi Chen, Omer
  Levy, Mike Lewis, Luke Zettlemoyer, and Veselin Stoyanov.
\newblock Roberta: A robustly optimized bert pretraining approach, 2019.

\bibitem[MacWhinney and Wagner(2010)]{macwhinney_wagner_2010}
Brian MacWhinney and Johannes Wagner.
\newblock Transcribing, searching and data sharing: The clan software and the
  talkbank data repository, 2010.
\newblock URL \url{https://www.ncbi.nlm.nih.gov/pmc/articles/PMC4257135/}.

\bibitem[Marmar et~al.(2019)Marmar, Brown, Qian, Laska, Siegel, Li,
  Abu‐Amara, Tsiartas, Richey, Smith, Knoth, and
  Vergyri]{marmar_speech-based_2019}
Charles~R. Marmar, Adam~D. Brown, Meng Qian, Eugene Laska, Carole Siegel, Meng
  Li, Duna Abu‐Amara, Andreas Tsiartas, Colleen Richey, Jennifer Smith, Bruce
  Knoth, and Dimitra Vergyri.
\newblock Speech-based markers for posttraumatic stress disorder in {US}
  veterans.
\newblock \emph{Depression and Anxiety}, 36\penalty0 (7):\penalty0 607--616,
  2019.
\newblock ISSN 1520-6394.
\newblock \doi{10.1002/da.22890}.
\newblock URL \url{https://onlinelibrary.wiley.com/doi/abs/10.1002/da.22890}.
\newblock \_eprint: https://onlinelibrary.wiley.com/doi/pdf/10.1002/da.22890.

\bibitem[Martin et~al.(2019)Martin, Muller, Suárez, Dupont, Romary, Éric
  Villemonte de~la Clergerie, Seddah, and Sagot]{martin2019camembert}
Louis Martin, Benjamin Muller, Pedro Javier~Ortiz Suárez, Yoann Dupont,
  Laurent Romary, Éric Villemonte de~la Clergerie, Djamé Seddah, and Benoît
  Sagot.
\newblock Camembert: a tasty french language model, 2019.

\bibitem[Mikolov et~al.(2013)Mikolov, Chen, Corrado, and
  Dean]{mikolov_efficient_2013}
Tomas Mikolov, Kai Chen, Greg Corrado, and Jeffrey Dean.
\newblock Efficient {Estimation} of {Word} {Representations} in {Vector}
  {Space}.
\newblock \emph{arXiv:1301.3781 [cs]}, September 2013.
\newblock URL \url{http://arxiv.org/abs/1301.3781}.
\newblock arXiv: 1301.3781.

\bibitem[Nozza et~al.(2020)Nozza, Bianchi, and Hovy]{nozza_what_2020}
Debora Nozza, Federico Bianchi, and Dirk Hovy.
\newblock What the [{MASK}]? {Making} {Sense} of {Language}-{Specific} {BERT}
  {Models}.
\newblock \emph{arXiv:2003.02912 [cs]}, March 2020.
\newblock URL \url{http://arxiv.org/abs/2003.02912}.
\newblock arXiv: 2003.02912.

\bibitem[Ortiz~Su{\'a}rez et~al.(2020)Ortiz~Su{\'a}rez, Romary, and
  Sagot]{ortiz-suarez-etal-2020-monolingual}
Pedro~Javier Ortiz~Su{\'a}rez, Laurent Romary, and Beno{\^\i}t Sagot.
\newblock A monolingual approach to contextualized word embeddings for
  mid-resource languages.
\newblock In \emph{Proceedings of the 58th Annual Meeting of the Association
  for Computational Linguistics}, pages 1703--1714, Online, July 2020.
  Association for Computational Linguistics.
\newblock URL \url{https://www.aclweb.org/anthology/2020.acl-main.156}.

\bibitem[Patel et~al.(2018)Patel, Saxena, Lund, Thornicroft, Baingana, Bolton,
  Chisholm, Collins, Cooper, Eaton, Herrman, Herzallah, Huang, Jordans,
  Kleinman, Medina-Mora, Morgan, Niaz, Omigbodun, Prince, Rahman, Saraceno,
  Sarkar, De~Silva, Singh, Stein, Sunkel, and UnÜtzer]{patel_lancet_2018}
Vikram Patel, Shekhar Saxena, Crick Lund, Graham Thornicroft, Florence
  Baingana, Paul Bolton, Dan Chisholm, Pamela~Y. Collins, Janice~L. Cooper,
  Julian Eaton, Helen Herrman, Mohammad~M. Herzallah, Yueqin Huang, Mark J.~D.
  Jordans, Arthur Kleinman, Maria~Elena Medina-Mora, Ellen Morgan, Unaiza Niaz,
  Olayinka Omigbodun, Martin Prince, Atif Rahman, Benedetto Saraceno, Bidyut~K.
  Sarkar, Mary De~Silva, Ilina Singh, Dan~J. Stein, Charlene Sunkel, and
  JÜrgen UnÜtzer.
\newblock The {Lancet} {Commission} on global mental health and sustainable
  development.
\newblock \emph{Lancet (London, England)}, 392\penalty0 (10157):\penalty0
  1553--1598, 2018.
\newblock ISSN 1474-547X.
\newblock \doi{10.1016/S0140-6736(18)31612-X}.

\bibitem[Smith(2017)]{smith2017cyclical}
Leslie~N Smith.
\newblock Cyclical learning rates for training neural networks.
\newblock In \emph{2017 IEEE Winter Conference on Applications of Computer
  Vision (WACV)}, pages 464--472. IEEE, 2017.

\bibitem[Su et~al.(2011)Su, Tsai, Hung, and Chou]{su_change_2011}
Jian-An Su, Ching-Shu Tsai, Tai-Hsin Hung, and Shih-Yong Chou.
\newblock Change in accuracy of recognizing psychiatric disorders by
  non-psychiatric physicians: five-year data from a psychiatric
  consultation-liaison service.
\newblock \emph{Psychiatry and Clinical Neurosciences}, 65\penalty0
  (7):\penalty0 618--623, December 2011.
\newblock ISSN 1440-1819.
\newblock \doi{10.1111/j.1440-1819.2011.02272.x}.

\bibitem[Thibaut(2018)]{thibaut_controversies_2018}
Florence Thibaut.
\newblock Controversies in psychiatry.
\newblock \emph{Dialogues in Clinical Neuroscience}, 20\penalty0 (3):\penalty0
  151--152, September 2018.
\newblock ISSN 1294-8322.
\newblock URL \url{https://www.ncbi.nlm.nih.gov/pmc/articles/PMC6296394/}.

\bibitem[Trevino et~al.(2011)Trevino, Quatieri, and
  Malyska]{trevino_phonologically-based_2011}
Andrea~Carolina Trevino, Thomas~Francis Quatieri, and Nicolas Malyska.
\newblock Phonologically-based biomarkers for major depressive disorder.
\newblock \emph{EURASIP Journal on Advances in Signal Processing},
  2011\penalty0 (1):\penalty0 42, August 2011.
\newblock ISSN 1687-6180.
\newblock \doi{10.1186/1687-6180-2011-42}.
\newblock URL \url{https://doi.org/10.1186/1687-6180-2011-42}.

\bibitem[Vaswani et~al.(2017)Vaswani, Shazeer, Parmar, Uszkoreit, Jones, Gomez,
  Kaiser, and Polosukhin]{vaswani2017attention}
Ashish Vaswani, Noam Shazeer, Niki Parmar, Jakob Uszkoreit, Llion Jones,
  Aidan~N. Gomez, Lukasz Kaiser, and Illia Polosukhin.
\newblock Attention is all you need, 2017.

\bibitem[Virtanen et~al.(2019)Virtanen, Kanerva, Ilo, Luoma, Luotolahti,
  Salakoski, Ginter, and Pyysalo]{virtanen2019multilingual}
Antti Virtanen, Jenna Kanerva, Rami Ilo, Jouni Luoma, Juhani Luotolahti, Tapio
  Salakoski, Filip Ginter, and Sampo Pyysalo.
\newblock Multilingual is not enough: Bert for finnish, 2019.

\bibitem[Wang et~al.(2018)Wang, Singh, Michael, Hill, Levy, and
  Bowman]{wang2018glue}
Alex Wang, Amanpreet Singh, Julian Michael, Felix Hill, Omer Levy, and Samuel~R
  Bowman.
\newblock Glue: A multi-task benchmark and analysis platform for natural
  language understanding.
\newblock \emph{arXiv preprint arXiv:1804.07461}, 2018.

\bibitem[Wolf et~al.(2019)Wolf, Debut, Sanh, Chaumond, Delangue, Moi, Cistac,
  Rault, Louf, Funtowicz, and Brew]{Wolf2019HuggingFacesTS}
Thomas Wolf, Lysandre Debut, Victor Sanh, Julien Chaumond, Clement Delangue,
  Anthony Moi, Pierric Cistac, Tim Rault, R'emi Louf, Morgan Funtowicz, and
  Jamie Brew.
\newblock Huggingface's transformers: State-of-the-art natural language
  processing.
\newblock \emph{ArXiv}, abs/1910.03771, 2019.

\bibitem[Xiong et~al.(2016)Xiong, Droppo, Huang, Seide, Seltzer, Stolcke, Yu,
  and Zweig]{xiong2016achieving}
Wayne Xiong, Jasha Droppo, Xuedong Huang, Frank Seide, Mike Seltzer, Andreas
  Stolcke, Dong Yu, and Geoffrey Zweig.
\newblock Achieving human parity in conversational speech recognition.
\newblock \emph{arXiv preprint arXiv:1610.05256}, 2016.

\bibitem[Yang et~al.(2020)Yang, Dai, Yang, Carbonell, Salakhutdinov, and
  Le]{yang_xlnet_2020}
Zhilin Yang, Zihang Dai, Yiming Yang, Jaime Carbonell, Ruslan Salakhutdinov,
  and Quoc~V. Le.
\newblock {XLNet}: {Generalized} {Autoregressive} {Pretraining} for {Language}
  {Understanding}.
\newblock \emph{arXiv:1906.08237 [cs]}, January 2020.
\newblock URL \url{http://arxiv.org/abs/1906.08237}.
\newblock arXiv: 1906.08237.

\end{thebibliography}

\end{document}